\documentclass{article}

\usepackage[position,preprint]{neurips_2026}

\usepackage[utf8]{inputenc}
\usepackage[T1]{fontenc}
\usepackage{hyperref}
\usepackage{url}
\usepackage{booktabs}
\usepackage{amsfonts}
\usepackage{nicefrac}
\usepackage{microtype}
\usepackage{xcolor}
\usepackage{amsmath} 
\usepackage{graphicx}

\hypersetup{
  hidelinks,
  pdftitle={World Models Are the Right Foundation for Simulating Mars Terraforming},
  pdfauthor={Sameera S Kashyap, Akshitha Nagaraj, Razvan Marinescu}
}
 
\title{World Models Are the Right Foundation for Simulating Mars Terraforming}

\author{
  Sameera S Kashyap \quad Akshitha Nagaraj \quad Razvan Marinescu \\
  Department of Computer Science and Engineering \\
  University of California, Santa Cruz \\
  \texttt{ssudars1@ucsc.edu, aknagara@ucsc.edu, ramarine@ucsc.edu}
}

\begin{document}

\maketitle

\begin{abstract}
Terraforming Mars is a long-horizon, action-conditioned control problem that requires predicting atmospheric responses to interventions over centuries to identify effective policies. This position paper argues that physics-regularized recurrent state-space world models are the most promising foundation for this setting. Data-driven climate emulators lack action-conditioning, and physics-based general circulation models (GCMs) are too expensive for policy search. Among learned world models, recurrent state-space architectures with physics constraints are better suited than transformer-based alternatives: they provide native uncertainty quantification through their stochastic-deterministic decomposition, remain stable under sparse non-terrestrial training data, and enforce physical consistency at each rollout step. To make this position concrete, we describe a potential architecture combining JEPA-inspired latent encoding, universal differential equation constraints, and a Dreamer-lineage planning head. We also introduce MarsWorld, an open benchmark proposal for intervention-conditioned planetary world models, and discuss broader implications and objections.
\end{abstract}

\section{Introduction}

On April 1, 2026, NASA's Artemis II mission completed the first crewed journey beyond low Earth orbit since Apollo 17, laying the groundwork for future crewed missions to Mars \citep{nasa2026artemis}. Humanity's long-term future may depend on extending habitable environments beyond Earth, and Mars is the most viable candidate for human expansion \citep{debenedictis2025mars}. Scientists and engineers are actively revisiting what it would take to make it habitable \citep{stork2025greenMars}, which raises a central open question: how do we simulate, plan for, and ultimately engineer the atmosphere of another planet?

Recent analysis demonstrates that human-relevant atmospheric pressures on Mars would require exaton-class atmospheric inventories, with accessible endogenous \(\mathrm{CO_2}\) delivering at most tens of millibars of warming. This is far short of habitability, implying that active engineering interventions spanning centuries to millennia will be necessary \citep{turyshev2026terraforming}. The engineering challenge is therefore not \emph{whether} to simulate Mars's atmosphere, but \emph{how} to do so fast enough and at sufficient scale to explore the space of possible intervention strategies.

A default run of the NASA Ames Legacy Mars global climate model simulates only 20 sols and takes about an hour, underscoring why large-scale policy search over terraforming strategies remains computationally prohibitive \citep{haberle2019mgcm}. Meaningful policy search, therefore, requires a faster simulation framework that makes this ensemble exploration computationally feasible.

Alongside these developments, the machine learning community has rapidly matured learned world model systems that compress environment dynamics into latent representations for planning through imagined experience. DreamerV3 \citep{hafner2025dreamerv3} showed that a single world model architecture can scale across many tasks, \citet{lecun2022path} established a principled latent-space prediction framework through JEPA, and Meta's V-JEPA~2 \citep{assran2025vjepa2} demonstrated that representation-space pretraining can transfer to physical robot control. Critically, \citet{wu2022daydreamer} showed that DreamerV3-style world models can be trained on real physical robots outside simulation, providing evidence that latent world model architectures transfer from game environments to physical systems governed by real dynamics, albeit at a scale far below planetary simulation. Yet these systems remain trained on terrestrial data and are not designed for non-terrestrial atmospheres or deliberate intervention over multi-century horizons.

This paper argues that learned latent \textbf{world models are the right foundation for simulating Mars terraforming}. The machine learning community already possesses much of the conceptual and architectural machinery needed to address this problem, but has not yet pointed it at Mars. Learned latent world models, specifically recurrent state-space architectures regularized by physics-informed priors \citep{hafner2019planet}, can reduce century-scale atmospheric rollouts by potentially orders of magnitude. This would enable reinforcement learning agents to plan across intervention sequences that no existing system can explore. We argue that building such a model is a concrete step toward ML systems capable of reasoning about planetary-scale interventions, and that the ML community should treat planetary engineering as a first-class research domain. The following sections establish the simulation gap (\S\ref{sec:gap}), propose the architectural path forward (\S\ref{sec:framework}), introduce a hypothetical benchmark (\S\ref{sec:benchmark}), and discuss broader implications and objections (\S\ref{sec:broader implications}, \S\ref{sec:alternatives}).

\section{Background}

\label{sec:background}
Terraforming Mars is not a single action but a multi-century control problem. \citet{mckay1991habitable} first established the thermodynamic requirements for habitability, and \citet{zubrin1997technological} quantified the positive CO$_2$ feedback dynamics that make staged intervention feasible in principle. Contemporary analysis suggests that Mars terraforming may be physically plausible in principle, but current simulation and planning tools are not sufficient to design a viable path: \citet{jakosky2018co2} established that accessible endogenous CO$_2$ reservoirs deliver at most tens of millibars of warming, far short of habitability, requiring active multi-century engineering. \citet{debenedictis2025mars} identifies simulation infrastructure as the critical near-term research gap.

\subsection{Mars general circulation models (M-GCM)}
Three general circulation models dominate Mars atmospheric simulation. MarsWRF \citep{richardson2007marswrf} solves the primitive equations of atmospheric dynamics on a structured latitude-longitude grid, including radiation transfer, dust transport, and CO$_2$ condensation. The Laboratory of Dynamical Meteorology Planetary Climate Model (LMD PCM) \citep{forget1999lmd} provides the European counterpart and has been validated against Thermal Emission Spectrometer (TES) and Mars Climate Sounder (MCS) observations over multiple Mars years. The NASA Ames M-GCM \citep{haberle2019mgcm} documents legacy physics packages and supports modular community use. Together, these models form the basis for Mars atmospheric simulation, and are validated against observational records spanning more than two decades \citep{greybush2019emars}. Gaussian-process emulation is useful for accelerating Mars GCM studies, but it does not fully solve the problem because it scales poorly with data size and is less reliable for strong perturbations or poorly covered regions of parameter space \citep{tunnell2023fast}. Ensemble Mars Atmosphere Reanalysis System (EMARS) \citep{greybush2019emars} addresses this by assimilating TES and MCS retrievals into a Mars atmospheric reanalysis, providing the closest analog to the ECMWF Reanalysis v5 (ERA5)-style data used to train Earth climate models.

\subsection{Machine Learning based atmospheric simulation for Earth}

The EarthML climate community has shown that neural dynamics models can simulate planetary-scale atmospheric physics at substantially lower cost than traditional GCMs. \citet{kochkov2024neuralgcm} introduced NeuralGCM, combining a differentiable dynamical core with learned physical parameterizations to achieve stable multi-decadal rollouts. \citet{bodnar2024aurora} trained Aurora on over one million hours of assimilated atmospheric data, producing a foundation model for Earth atmospheric prediction. \citet{cresswellclay2025dlesym} demonstrated century-scale coupled atmosphere ocean simulation with Deep Learning Earth System Model (DLESyM), while \citet{wattmeyer2024ace2} showed that ACE2 can learn subseasonal to decadal variability and forced responses with physical consistency. \citet{vankatwyk2026rewiring} argue that the next phase of climate modeling will depend on tighter co-design between physics-based simulators and learned ML emulators. These systems are all trained on ERA5 or similar Earth reanalysis data, and none have been applied to a non-terrestrial planetary atmosphere.

\subsection{Climate Modeling with Physics-Informed Neural Networks}

Physical constraints are essential for Mars terraforming simulation because the model must operate far outside the training distribution, in pressure, composition, and temperature regimes with no historical precedent, where a purely data-driven model has no mechanism to prevent physically impossible predictions. Physics-informed neural networks (PINNs) address this by embedding governing differential equations into the loss function, encouraging physically consistent predictions under distribution shift \citep{raissi2019pinns, cuomo2022scientific}. Universal Differential Equations (UDEs) \citep{rackauckas2020ude} extend this idea by embedding neural networks inside ODE solvers: known large-scale atmospheric dynamics are encoded analytically, while poorly characterized Martian processes such as dust storms, polar vortex dynamics, and CO$_2$ ice microphysics are learned as residual terms. \citet{kashyap2024sciml} showed that this approach can improve accuracy on chaotic atmospheric systems under reduced data, which is directly relevant to Mars's sparse observational record. Recent work on continuous-time atmospheric modeling provides further support: \citet{verma2024climode} introduced ClimODE, a neural ODE that learns atmospheric advection dynamics with fewer parameters than standard data-driven baselines, and \citet{rogers2023exoplanet} demonstrated that physics-informed neural emulators can accelerate planetary atmospheric evolution by orders of magnitude while remaining close to the original physics model.

\subsection{World Models simulate physical environments}

World models learn compressed latent representations of environment dynamics and train agents through imagined rollouts rather than direct interaction with the environment \citep{ha2018worldmodels}. The Recurrent State-Space Model (RSSM) \citep{hafner2019planet} introduced the stochastic-deterministic latent backbone that underlies many modern world models, and DreamerV3 \citep{hafner2025dreamerv3} showed that this approach can scale across a wide range of tasks. For continuous control, Temporal Difference Model Predictive Control 2 (TD-MPC2) \citep{hansen2024tdmpc2} demonstrated stable policy optimization in high-dimensional continuous state spaces, while Recall to Imagine (R2I) \citep{samsami2024r2i} extended Dreamer-style memory for longer-horizon credit assignment. Together, these works establish latent world models as a practical framework for sequential prediction and control.

A parallel line of work has focused on predicting abstract representations rather than reconstructing pixels. JEPA \citep{lecun2022path} proposed latent-space prediction as an alternative to pixel-level generation, and I-JEPA \citep{assran2023ijepa}, V-JEPA 2 \citep{assran2025vjepa2}, and LeWorldModel \citep{maes2026leworldmodel} further demonstrated that representation-space models can scale and remain stable under limited compute and data. More recently, generative world models such as Genie 2 \citep{parkerholder2024genie2} and NVIDIA Cosmos \citep{nvidia2025cosmos} showed that world-model-style systems can generate interactive environments, but are optimized for visual realism rather than physically grounded atmospheric prediction. That limitation motivates a Mars-specific world model designed for numerical accuracy, intervention conditioning, and long-horizon stability.

\section{The non-intersection of climate simulators and World Models}
\label{sec:gap}

The core gap is that no existing system satisfies the full set of requirements for Mars terraforming simulation: \emph{(i)} physical accuracy under Martian atmospheric constraints, \emph{(ii)} computational tractability for ensemble policy search, \emph{(iii)} action-conditioned rollouts, \emph{(iv)} stable multi-century latent dynamics, and \emph{(v)} trainability under sparse non-terrestrial observational data. Table~\ref{tab:comparison} summarizes this mismatch. 

\begin{table}[htbp]
\centering
\footnotesize
\caption{Requirements for Mars terraforming simulation vs.\ 
existing systems. \textbf{(i)}~Mars physical accuracy; 
\textbf{(ii)}~tractable for policy search; 
\textbf{(iii)}~action-conditioned rollouts; 
\textbf{(iv)}~century-scale stability; 
\textbf{(v)}~trainable on sparse non-terrestrial data.}
\label{tab:comparison}
\begin{tabular}{lccccc}
\toprule
\textbf{System} & \textbf{i} & \textbf{ii} & 
  \textbf{iii} & \textbf{iv} & \textbf{v} \\
\midrule
\multicolumn{6}{l}{\textit{\textbf{Physics-based simulators}}} \\[2pt]
MarsWRF & \checkmark & \texttimes & \texttimes & \texttimes & \checkmark \\ LMD PCM   & \checkmark & \texttimes & \texttimes & \texttimes & \checkmark \\

\multicolumn{6}{l}{\textit{Earth ML climate models}} \\[2pt]
NeuralGCM           & \texttimes & \checkmark & \texttimes & \texttimes & \texttimes \\
ACE2                & \texttimes & \checkmark & \texttimes & \checkmark & \texttimes \\
Aurora              & \texttimes & \checkmark & \texttimes & \texttimes & \texttimes \\
DLESyM              & \texttimes & \checkmark & \texttimes & \checkmark & \texttimes \\

\multicolumn{6}{l}{\textit{World model architectures}} \\[2pt]
DreamerV3           & \texttimes & \checkmark & \checkmark & \texttimes & \texttimes \\
Genie 2/3           & \texttimes & \texttimes & \checkmark & \texttimes & \texttimes \\
\multicolumn{6}{l}{\textit{Mars-specific ML}} \\[2pt]
MarsGen             & \checkmark & \checkmark & \texttimes & \texttimes & \checkmark \\
\multicolumn{6}{l}{\textit{Physics-constrained ML}} \\[2pt]
PINNs / UDEs        & \checkmark & \texttimes & \texttimes & \texttimes & \checkmark \\
\midrule
\textbf{(Proposed) Mars World Model} 
  & \checkmark & \checkmark & \checkmark & \checkmark & \checkmark \\
\bottomrule
\end{tabular}
\end{table}

Earth ML climate models such as NeuralGCM \citep{kochkov2024neuralgcm}, ACE2 \citep{wattmeyer2024ace2}, Aurora \citep{bodnar2024aurora}, and DLESyM \citep{cresswellclay2025dlesym} provide tractable rollouts on Earth data, but do not satisfy Martian physical accuracy or action-conditioned planning. Physics-based Mars simulators such as MarsWRF \citep{richardson2007marswrf} and the LMD PCM \citep{forget1999lmd} satisfy Martian physics by construction, but are too expensive for policy search and do not accept sequential intervention actions. World model architectures such as DreamerV3 and Genie solve the action-conditioned planning problem, but are not designed for Martian physics or sparse non-terrestrial training data. MarsGen \citep{li2025marsgen} is a Mars-specific generative model that produces realistic visual sequences of Martian surface and atmospheric dynamics, but it lacks action-conditioning and century-scale rollout stability.  

Terraforming simulations of Mars are especially challenging because they combine non-stationary dynamics under intervention and sparse observational coverage; Mars’s CO$_2$-dominated, low-pressure, dust-driven atmosphere undergoes seasonal phase transitions with no Earth analogue, so terraforming pushes the system away from its historical distribution at every step. EMARS \citep{greybush2019emars} provides the relevant observational foundation, but spans only ten Mars years of assimilated fields, far too sparse to train a data-driven model without strong physical priors. This motivates a physics-informed recurrent state-space world model that combines physical constraints, tractable latent rollouts, and action-conditioned planning.

\section{A framework for planetary atmospheric World Models}
\label{sec:framework}

\begin{figure}[htbp]
\centering
\includegraphics[width=0.85\columnwidth]{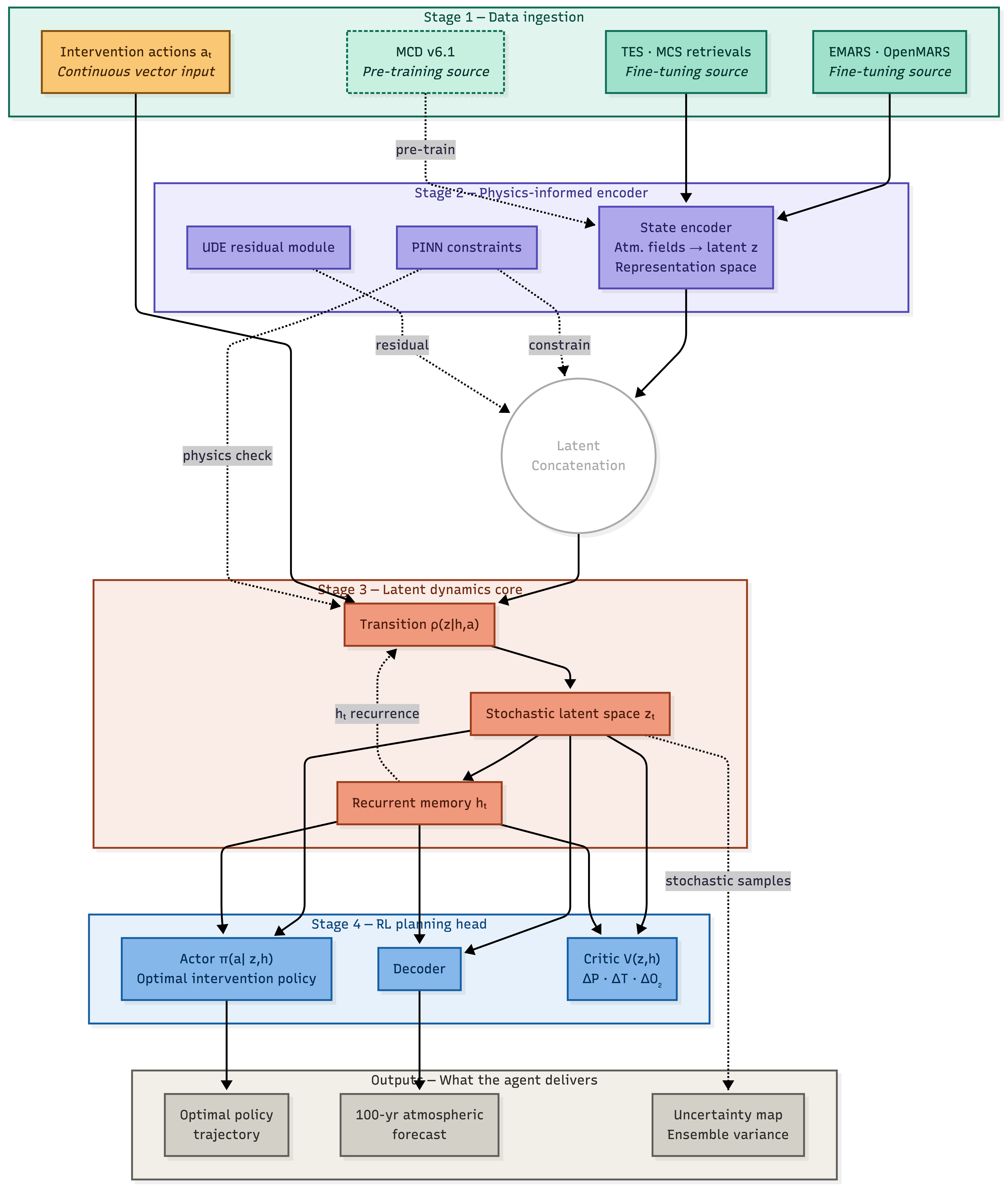}
\caption{Proposed Mars atmospheric world model architecture.
\textbf{S1}: A two-stage training strategy separates synthetic pre-training data (MCD v6.1) from observational fine-tuning sources (EMARS, OpenMARS; with TES/MCS as the underlying retrievals); intervention actions $a_t$ enter the transition model as continuous vectors. \textbf{S2}: A state encoder maps atmospheric fields to a latent $z$; PINN constraints and UDE residuals regularize the latent and downstream dynamics, merged at the concatenation node. \textbf{S3}: An RSSM propagates stochastic latent $z_t$ and recurrent memory $h_t$ under action-conditioned transitions $\rho(z \mid h, a)$ with physics enforced at every step; $h_t$ is updated from $(h_{t-1}, z_{t-1}, a_{t-1})$ via the recurrent backbone. \textbf{S4}: An actor-critic head trains within imagined latent rollouts; the actor $\pi(a \mid z, h)$ outputs intervention policies, the critic $V(z, h)$ estimates returns from $(\Delta P, \Delta T, \Delta O_2)$, and a decoder reconstructs gridded fields against EMARS as an auxiliary signal. \textbf{S5}: Outputs are a learned policy trajectory, a long-horizon atmospheric forecast, and a rollout uncertainty map from stochastic latent samples.}
\label{fig:architecture}
\end{figure}

We propose a four-stage architecture for a Mars atmospheric world model that synthesizes the strengths of the JEPA representation learning lineage, the DreamerV3 latent dynamics framework, physics-informed neural network constraints, and Universal Differential Equation residual modeling. The architecture is designed to satisfy all 
five criteria identified in Section~\ref{sec:gap} 
simultaneously. Figure~\ref{fig:architecture} illustrates the full pipeline.

\subsection{Stage 1: Data ingestion and preprocessing}

The primary observational foundation for the proposed model is the Ensemble Mars Atmosphere Reanalysis System (EMARS) \citep{greybush2019emars} and the Open Access to Mars Assimilated Remote Soundings (OpenMARS) dataset \citep{holmes2020openmars}, which provide gridded fields of temperature, pressure, dust optical depth, wind, CO$_2$ surface ice, and water ice spanning over Mars Year 24 to Mars Year 33. These reanalyses are assimilated from Thermal Emission Spectrometer (TES) \citep{christensen2001tes} and Mars Climate Sounder (MCS) \citep{mccleese2007mcs} retrievals using ensemble data assimilation techniques, making them the closest Martian counterpart to the ERA5 reanalysis used for Earth-system ML. To address the sparsity of this observational record relative to ERA5's multidecadal coverage, we propose augmenting training with synthetic rollouts from the Mars Climate Database (MCD v6.1) \citep{millour2022mcd61}. The MCD enables sampling a wide range of Martian environmental conditions, including dust-cycle intensity, seasonal phase coverage, and solar activity, which are less constrained in the observational record. We therefore propose pretraining on MCD synthetic scenarios to learn robust physical priors and then fine-tune on EMARS and OpenMARS fields, following the broader emulator-based strategy used by \citep{rogers2023exoplanet} for accelerating planetary atmosphere inference. Intervention actions, including greenhouse gas injection, polar albedo modification, and orbital mirror insolation, are encoded as continuous vectors that condition the transition model at each timestep, following the action-conditioned design of \citep{hafner2019planet}.

\subsection{Stage 2: Physics-informed encoder}

The encoder compresses the high-dimensional Martian atmospheric state into a compact latent representation that is regularized by physical constraints and supplemented by learned residual dynamics. The \emph{state encoder} maps gridded atmospheric fields $\mathbf{x}_t = \{T, P, \tau, \mathbf{u}\}$, where $T$ is temperature, $P$ is pressure, $\tau$ is dust optical depth, and $\mathbf{u}$ is the wind vector into a dense latent vector $z \in \mathbb{R}^d$, operating in abstract representation space rather than grid space following \citet{lecun2022path}. This design discards information irrelevant to dynamics while preserving the structure needed for planning, a choice that is critical given the limited observational record. \emph{PINN hard constraints} \citep{raissi2019pinns, cuomo2022scientific} regularize the latent representation and dynamics rollouts at every step through three physical loss terms: hydrostatic balance
\begin{equation}
\frac{\partial p}{\partial \zeta} = -\rho g,
\end{equation}
where $\zeta$ is altitude, $\rho$ is atmospheric density, and
$g$ is gravitational acceleration. This prevents unphysical vertical pressure gradients; global dry air mass conservation,
\begin{equation}
\int_{\text{col}} \rho \, d\zeta = \text{const},
\end{equation}
which prevents non-condensable atmospheric mass from being created or destroyed during rollout; and the CO$_2$ condensation bound,
\begin{equation}
T_s \geq T_{\text{CO}_2}(p_s),
\end{equation}
where $T_s$ is surface temperature, $p_s$ is surface pressure, and $T_{\text{CO}_2}(p_s)$ is the CO$_2$ frost point temperature at surface pressure. This prevents surface temperatures from falling below the condensation point of the dominant atmospheric constituent. A \emph{UDE residual module} \citep{rackauckas2020ude} represents poorly characterized Martian-specific processes such as dust storm nucleation, polar vortex dynamics, and perchlorate photochemistry as a neural network $\mathcal{N}_\theta$ embedded inside the dynamics ODE solver,
\begin{equation}
\frac{d\mathbf{x}}{dt} = f_{\text{phys}}(\mathbf{x}, t) + \mathcal{N}_\theta(\mathbf{x}, t),
\end{equation}
where $f_{\text{phys}}$ encodes known atmospheric dynamics analytically and $\mathcal{N}_\theta$ learns the residual from EMARS observations. The state encoder produces the latent input to the dynamics core, while PINN constraints and UDE residuals are applied at every step of the latent rollout in Stage~3.

\subsection{Stage 3: Recurrent state-space latent 
dynamics core}

The latent dynamics model propagates the encoder output forward in time through two coupled latent states and a learned transition. The \emph{stochastic latent vector} $z_t \in \mathbb{R}^d$ is drawn from a learned distribution $p_\theta(z_t \mid h_t, a_t)$ that we suggest parameterizing as a vector of independent categorical variables with straight-through gradient estimation, following the discrete latent formulation of DreamerV3 \citep{hafner2025dreamerv3}. This choice provides better long-horizon stability than diagonal Gaussian latents under century-scale rollouts. At each step, $z_t$ encodes a sampled compressed state of the atmosphere consistent with the temperature, pressure, dust, and wind fields ingested by the encoder. The \emph{recurrent memory} $h_t$, updated as
\begin{equation}
h_t = \text{Recurrent}(h_{t-1}, z_{t-1}, a_{t-1}),
\end{equation}
using either a GRU or an S4-style structured state-space module \citep{samsami2024r2i}, maintains a compressed causal history of prior atmospheric states and intervention actions across long horizons. These memory pathways preserve long-range dependencies that a standard recurrent unit may struggle to retain over century-scale simulated rollouts. The \emph{transition model} $p(z_t \mid h_t, a_t)$ then advances the latent state conditioned on the intervention action $a_t$, with PINN constraints applied at each prediction step to discourage rollouts that violate thermodynamic bounds and UDE residuals applied within the dynamics ODE solver to capture poorly characterized Martian processes. The recurrent feedback from $h_t$ to subsequent stochastic latents produces a coherent multi-step latent dynamical system over long planning horizons.

We recommend the use of RSSM rather than a transformer-based world model for three reasons. First, RSSM represents uncertainty explicitly through its stochastic latent state: $p_\theta(z_t \mid h_t, a_t)$ is a discrete categorical distribution over latent vectors $z_t$, from which we propose sampling multiple physically plausible atmospheric futures at each step, while $h_t$ preserves the deterministic history of prior states and interventions. This is essential for downstream policy search, where the agent must distinguish genuinely uncertain predictions from physically implausible ones. Second, RSSM handles data scarcity better. With roughly a decade in Mars years of observational data, a large transformer world model would likely overfit; structured RSSM transitions combined with physics-informed regularization remain stable under sparse observations. Third, transformer world models such as DIAMOND \citep{alonso2024diffusionworldmodelingvisual}, IRIS \citep{micheli2023transformerssampleefficientworldmodels}, and DreamerV4 \citep{hafner2025trainingagentsinsidescalable} scale well and handle long sequences, but are primarily optimized for pixel-level reconstruction rather than physically grounded state prediction with calibrated uncertainty. Transformer world models remain a natural direction for future work as the EMARS record expands and the data-scarcity argument weakens.

\subsection{Stage 4: Reinforcement Learning (RL) planning head}

The RL planning head trains an actor-critic policy within imagined latent rollouts, following the Dreamer paradigm \citep{hafner2019dreamerv1, hafner2025dreamerv3}. After world-model training, policy learning proceeds primarily in latent space, which is intended to substantially reduce the need for expensive GCM rollouts while still permitting long-horizon intervention evaluation. The \emph{decoder} reconstructs gridded atmospheric fields from $(z_t, h_t)$, providing an auxiliary reconstruction loss against EMARS during world-model training and keeping latent states physically interpretable. The \emph{actor} $\pi(a \mid z_t, h_t)$ outputs a continuous intervention vector specifying greenhouse gas injection rate, polar albedo modification, and orbital mirror insolation factor. The \emph{critic} $V(z_t, h_t)$ estimates expected cumulative reward over the remaining terraforming horizon, enabling optimization of long-term rather than immediate outcomes. The reward $R(z_t, h_t)$ combines three physically grounded habitability proxies: surface pressure increase $\Delta P$ toward the 0.611~kPa triple point of water and ultimately the $\sim$30~kPa Armstrong limit for unpressurized human survival, mean surface temperature $\Delta T$ toward 273~K, and assumed atmospheric oxygen fraction $\Delta O_2$ as a long-horizon biological viability indicator. Return targets use DreamerV3's symlog-transformed values and percentile-normalized returns \citep{hafner2025dreamerv3} to stabilize learning across the wide reward-scale variation in century-scale simulation. For continuous intervention spaces, we suggest using latent-space trajectory optimization in the spirit of TD-MPC2 \citep{hansen2024tdmpc2} as an optional planning refinement rather than a replacement for the actor-critic objective.

\section{The MarsWorld benchmark}
\label{sec:benchmark}
We introduce MarsWorld, a proposed open benchmark for evaluating planetary atmosphere world models under explicit intervention. Since evaluation trajectories require expensive multi-GCM rollouts, we present MarsWorld as a benchmark \emph{specification}: the task definition, metrics, and evaluation protocol are outlined here, with reference trajectory generation deferred to future work.

\subsection{Task definition}
Given the EMARS atmospheric state at Mars Year~24 \citep{greybush2019emars}, a model must predict atmospheric evolution over a decade of Mars years ($\sim$19 Earth years) under a controlled intervention sequence. We consider three intervention families drawn from the terraforming literature: \emph{(i)} greenhouse gas injection via perfluorocarbon release \citep{gerstell2001pfc}, \emph{(ii)} reduction of polar cap albedo \citep{mckay1991habitable}, and \emph{(iii)} increase of orbital insolation via deployed mirrors \citep{zubrin1997technological}. Each intervention contributes one or more dimensions to a continuous action vector $a_t \in \mathbb{R}^d$ applied at seasonal ($L_s$) granularity.

\subsection{Evaluation protocol and metrics}
To avoid the simulator-leakage concern discussed in Section~\ref{sec:alternatives}, evaluation trajectories are generated by an independent GCM family rather than by LMD PCM, drawing on the multi-model precedent established by \citep{newman2021jezero}. Specifically, we propose held-out trajectories from MarsWRF \citep{richardson2007marswrf} and the NASA Ames MGCM \citep{haberle2019mgcm} under intervention sequences not present in pre-training data. MarsWorld reports four metrics separately, since models may fail along independent axes:

\begin{enumerate}
    \item \textbf{Intervention fidelity} (primary): per-variable normalized RMSE against held-out GCM trajectories for temperature, pressure, and dust optical depth at 1, 5, and 10 simulated Mars years. Reported separately by variable to avoid scale-aggregation ambiguity.
    \item \textbf{Physical consistency}: fraction of rollout timesteps satisfying hydrostatic balance (relative violation $< 1\%$), non-condensable dry air mass conservation (drift $<$ 5\% per Mars year), and the CO$_2$ condensation bound $T_s \geq T_{\mathrm{CO}_2}(p_s)$.
    \item \textbf{Rollout stability}: maximum rollout length before surface pressure leaves the envelope spanned by physics-based PCM simulations under the same intervention magnitude.
    \item \textbf{Closed-loop policy utility}: a learned policy is unrolled in the world model to obtain an intervention sequence, then executed in a held-out GCM as the test environment; we report the achieved habitability-relevant surface pressure $\Delta P$ at the 10-year horizon. This closed-loop protocol measures true policy transferability rather than optimism inside the model.
\end{enumerate}

\section{Broader implications}
\label{sec:broader implications}
The framework proposed in Section \ref{sec:framework} also raises open questions that extend beyond Mars. Physics-regularized recurrent world models trained on sparse, non-terrestrial data stress inductive bias, sample efficiency, and physical generalization in ways that terrestrial benchmarks do not. MarsWorld is primarily a planetary science benchmark, but it creates open ML problems under conditions that are unusually severe and physically interpretable. We highlight two where the benchmark contributes something not available in existing evaluation settings.

\textbf{Out-of-distribution generalization under extreme covariate shift.} Most OOD benchmarks involve moderate distribution shifts: a new image domain, a held-out region, or a different sensor. MarsWorld is categorically more extreme. A model trained on ten Mars years of EMARS must extrapolate across pressure increases of orders of magnitude, CO$_2$ phase transitions, and dust radiative forcing giving way to greenhouse gas forcing, where every intervention pushes the state further from the training distribution at every rollout step. \citet{navarro2026assessingrobustnessclimatefoundation} demonstrate that high-capacity climate foundation models degrade substantially under no-analog forcing regimes; Mars terraforming takes this failure mode to its logical extreme by design. MarsWorld provides a controlled setting for studying whether physics-informed inductive biases can substitute for distributional coverage when the latter is structurally unavailable, a question relevant well beyond planetary simulation.

\textbf{Long-horizon credit assignment in learned simulators.} Model-based RL benchmarks such as DMControl \citep{tassa2018dmcontrol}, Atari \citep{bellemare2013ale}, and Crafter \citep{hafner2022crafter} have driven substantial progress in world model planning, but their effective horizons remain on the order of hundreds to low thousands of environment steps. MarsWorld operates at a fundamentally different scale: habitability-relevant rewards such as surface pressure approaching the triple point of water and mean temperature approaching 273 K all accumulate over simulated decades, requiring credit assignment across far longer and sparser trajectories. Current latent world model architectures have not been demonstrated at this horizon; \citet{maes2026leworldmodel} explicitly note that long-horizon planning remains an open problem and identify hierarchical decomposition as a necessary research direction. MarsWorld makes this limitation empirically measurable: a model that achieves high intervention fidelity but low policy utility score reveals a gap between predictive accuracy and planning capability, surfacing the credit assignment problem as a concrete empirical finding rather than a theoretical concern.

\textbf{Causal World Modeling Under Intervention.} MarsWorld also provides a controlled setting for studying causal world modeling under intervention. Unlike observational benchmarks, it asks whether a model can attribute atmospheric changes to specific actions rather than fit passive correlations, though its causal interpretation remains bounded by the assumptions of the underlying physics simulator.

\section{Alternative views and counterarguments}
\label{sec:alternatives}

A rigorous position paper must engage with the strongest available versions of its objections. We address five concerns below. The first two are the most damaging to the benchmark design; the remaining three address the modeling framework and its relevance to the ML community.

\textbf{Simulator leakage: does MarsWorld measure generalization or interpolation?} If pre-training uses MCD v6.1 and evaluation uses LMD PCM rollouts, both belong to the same family of physics-based GCMs and share broad assumptions and overlapping parameterizations. A model could score well by interpolating between simulator modes rather than learning genuine atmospheric dynamics. MarsWorld, therefore, cannot by itself establish real-world out-of-distribution generalization; it is a controlled testbed for intervention-aware world models within a specified physical model class. The natural fixes are held-out forcing combinations, cross-validation against independent GCM families such as MarsWRF \citep{richardson2007marswrf} following the multi-model evaluation precedent established by \citep{newman2021jezero} while acknowledging that full simulator-independence is not claimable without intervention data from real Mars.

\textbf{Action identifiability: can the benchmark distinguish causal control from correlation?}
Greenhouse gas injection, polar albedo reduction, and orbital insolation increase can produce partially overlapping atmospheric trajectories, making it difficult to attribute pressure or temperature changes to any single intervention. The benchmark does not solve this identification problem; instead, it makes it visible by reporting per-intervention sensitivity for each habitability metric, in a causal-attribution spirit inspired by do-calculus \citep{pearl2009causality}. Scalar aggregate scores should not be treated as sufficient. Some intervention pairs will remain confounded, and that limitation should be treated as open.

\textbf{Why a learned world model and not a cheaper surrogate?}
A more precise objection is why policy search needs a learned latent world model rather than a Gaussian process emulator \citep{tunnell2023fast}, polynomial chaos surrogate, or reduced-order PDE model. Those approaches are simpler and useful for fixed-boundary prediction, but they do not accept sequential actions and propagate their effects through the latent state over long horizons. That action-conditioning requirement is what motivates a recurrent world model. The claim is not that world models are universally superior, but that they are the natural architecture for sequential terraforming policy optimization.

\textbf{Does physics-informed inductive bias actually reduce sample complexity?}
This is plausible, but not yet demonstrated here. Physics constraints shrink the hypothesis space by excluding physically invalid models, and UDE residuals reduce the learning burden by handling the known large-scale dynamics analytically \citep{raissi2019pinns, rackauckas2020ude}. Whether that is sufficient for stable Mars forecasting with limited data is exactly what MarsWorld should test. Strong sample-complexity claims should wait for ablations against data-only and physics-only baselines.

\textbf{Is this ML research or application-driven engineering?}
This is a legitimate concern, but MarsWorld instantiates open ML problems rather than replacing them with domain-specific engineering. In particular, it stresses OOD generalisation under interventional covariate shift, long-horizon credit assignment with sparse rewards, and causal world modelling under sequential action. These are conditions under which current world models often struggle. The ML contribution should therefore be judged by whether the approach transfers beyond MarsWorld to other long-horizon causal planning settings.

\section{Conclusion}
We argued that learned latent world models, combining physical constraints, latent dynamics, and sequential control, are a promising foundation for Mars terraforming simulation. They fill a gap left by physics-only GCMs, which are too expensive for intervention search, climate models, which are not built for counterfactual action-conditioned regimes, and pixel-trained world models, which optimize perceptual fidelity rather than physically grounded atmospheric state. We sketched a PINN-constrained, RSSM-based world model with an actor-critic planning head, and established MarsWorld, an open benchmark proposal for evaluating intervention-aware planetary world models along the dimensions that matter: physical consistency, rollout stability, intervention fidelity, planning utility, and computational efficiency. We hope MarsWorld encourages the ML community to treat planetary engineering as a first-class research domain.

% ---- Bibliography ----
\bibliographystyle{abbrvnat}
\bibliography{references}

\end{document}